\documentclass{article}

\usepackage[preprint]{neurips_2025}

\usepackage{xspace}
\usepackage{tcolorbox}
\usepackage[utf8]{inputenc} %
\usepackage[T1]{fontenc}    %
\usepackage{hyperref}       %
\usepackage{url}            %
\usepackage{booktabs}       %
\usepackage{amsfonts}       %
\usepackage{nicefrac}       %
\usepackage{microtype}      %
\usepackage{xcolor}         %
\usepackage{amsmath}
\usepackage{bbm}
\usepackage{array}

\usepackage{cleveref}
\usepackage{multirow}
\usepackage{colortbl}
\usepackage{siunitx}
\usepackage{enumitem}
\usepackage{adjustbox}
\usepackage{xfrac}

\usepackage{soul} %
\usepackage{xcolor} %

\definecolor{verylightgray}{rgb}{0.9, 0.9, 0.9}
\definecolor{verylightblue}{rgb}{0.9, 0.95, 1.0}
\definecolor{verylightpurple}{rgb}{0.95, 0.9, 1.0}
\newcommand{\hlgray}[1]{\sethlcolor{verylightgray}\hl{#1}}

\title{Done Is Better than Perfect: Unlocking Efficient Reasoning by Structured Multi-Turn Decomposition}

\def\method{Multi-Turn Decomposition\xspace}
\def\abbr{MinD\xspace}

\author{%
Zihao Zeng\textsuperscript{1}\textsuperscript{2}\thanks{Equal contribution.}\;,
Xuyao Huang\textsuperscript{1}\footnotemark[1]\;,
Boxiu Li\textsuperscript{1},
Hao Zhang\textsuperscript{3},
and Zhijie Deng\textsuperscript{1}\thanks{Corresponding author.} \\
\textsuperscript{1}Shanghai Jiao Tong University~\textsuperscript{2}RealAI\\
\textsuperscript{3}University of California, San Diego\\
\texttt{\{zengzihao,\;huangxuyao,\;lbxhaixing154\}@sjtu.edu.cn}\\
\texttt{haozhang@ucsd.edu,\;zhijied@sjtu.edu.cn}
}

\begin{document}

\maketitle

\begin{abstract}

Large Reasoning Models (LRMs) have gained increasing attention over the past few months.
Despite being effective, LRMs are criticized for the excessively lengthy Chain-of-Thought (CoT) to derive the final answer, suffering from high first-token and overall latency.
Typically, the CoT of LRMs mixes multiple \emph{thinking units}, some of which are split by markers like ``aha'', ``wait'', or ``alternatively''; each unit attempts to produce a candidate answer to the original query.
Hence, a natural idea to improve efficiency is to reduce the unit number.
Yet, the fact that the {thinking units} in vanilla CoT cannot be explicitly managed renders doing so challenging.
This paper introduces \textbf{\method (\abbr)} to decode conventional CoT into a sequence of explicit, structured, and turn-wise interactions to bridge the gap.
In {\abbr}, the model provides a multi-turn response to the query, where each turn embraces a thinking unit and yields a corresponding answer.
The subsequent turns can reflect, verify, revise, or explore alternative approaches to both the thinking and answer parts of earlier ones.
This not only makes the answer delivered more swiftly,
but also enables explicit controls over the iterative reasoning process (i.e., users may halt or continue at any turn).
We follow a supervised fine-tuning (SFT) then reinforcement learning (RL) paradigm to realize {\abbr}.
We first rephrase the outputs of an
LRM into multi-turn formats by prompting another LLM, and then tune the LRM with such data.
Observing that the tuned model tends to consume even more tokens than the original one (probably due to that the multi-turn formats introduce additional answer tokens), we advocate leveraging RL algorithms like GRPO to prioritize correct outputs with fewer turns.
Trained on the MATH dataset using R1-Distill models, {\abbr} can achieve up to $\sim70\%$ reduction in both output token usage and time to first token (TTFT), while maintaining competitive performance on reasoning benchmarks such as MATH-500, AIME24, AMC23, and GPQA-Diamond.

\end{abstract}

\begin{figure}[ht]
    \centering
    \includegraphics[width=\linewidth]{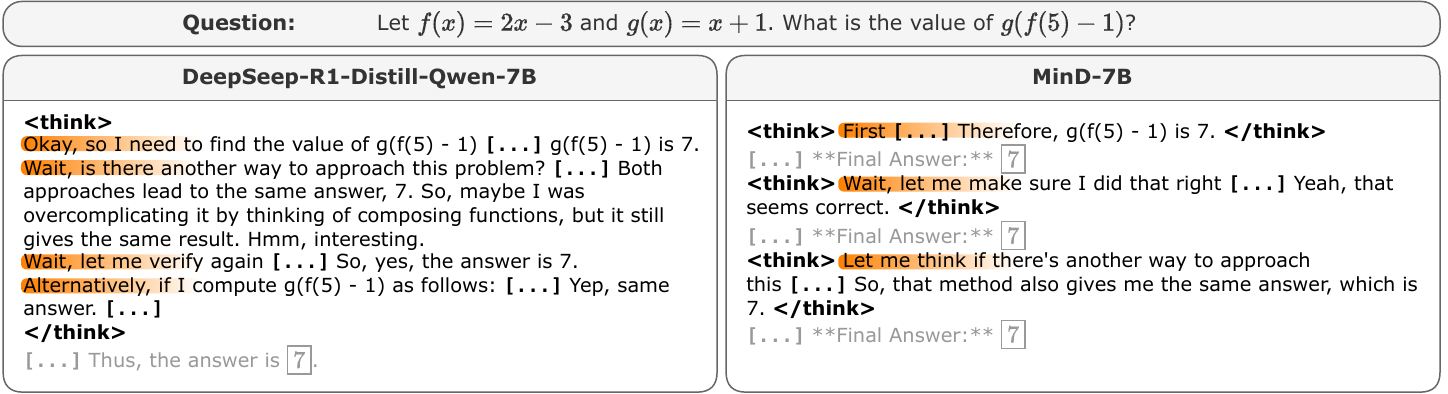}
    \caption{An illustration of responses from DeepSeek-R1-Distill-Qwen-7B and the transformed MinD-7B model on the same math problem. The original LRM follows a think-then-answer format, where the reasoning process consists of multiple thinking units (the start of each new unit is marked with an orange highlight). In contrast, MinD-7B adopts a multi-turn reasoning paradigm, where each turn contains a thinking unit followed by an answer. Also note that MinD-7B tends to use fewer thinking units due to the GRPO training (see Section~\ref{sec:method-mind}).}
    \label{fig:intro}
    \vspace{-10pt}
\end{figure}

\section{Introduction}
\label{sec:introduction}

Large Reasoning Models (LRMs) have recently attracted significant attention due to their advancing reasoning capabilities, including OpenAI-o1~\citep{jaech2024openai}, DeepSeek-R1~\citep{guo2025deepseek}, and Kimi-1.5~\citep{team2025kimi}.
These models have achieved remarkable performance on complex tasks, e.g., mathematical competitions, thanks to their ability to engage in a ``think-then-answer'' paradigm, where intermediate reasoning chains are generated to induce the final answer.
The resultant Chain-of-Thought (CoT) activates contextually accurate responses through iterative exploration and verification of potential solutions.

Despite these advantages, LRMs often suffer from inefficiency issues as the CoT can become excessively lengthy, exhibiting substantially increased computational costs and latency compared to non-reasoning Large Language Models (LLMs).
To mitigate these,
several strategies have been proposed in recent works.
For example, some approaches encourage models to generate answers more directly through strategically designed prompts~\citep{jie2024promptbasedlengthcontrolledgeneration}, truncate the chain of thought to avoid unnecessary token generation~\citep{fu2025reasoning,qwen3}, or leverage speculative reasoning via model collaboration~\citep{pan2025specreason, she2025hawkeyeefficientreasoningmodelcollaboration}.
Other approaches focus on reducing token redundancy by refining model reasoning paths through supervised fine-tuning (SFT)~\citep{yang2025thinkingoptimalscalingtesttimecompute}, or by enhancing decision efficiency with improvements to Group Relative Policy Optimization (GRPO) algorithms~\citep{yu2025dapoopensourcellmreinforcement,liu2025understandingr1zeroliketrainingcritical}.

The CoT reasoning process in LRMs is typically composed of multiple \emph{thinking units}---discrete cognitive steps like initial attempts, follow-up validations, reflections, and strategic shifts.
Each unit can contribute to generating a candidate answer, while current LRMs tend to employ redundant units to ensure the final answer is close to `perfect' (see an empirical analysis of such redundancy in \Cref{fig:method-1} (right)).
While reducing the number of thinking units could improve reasoning efficiency, the inability to explicitly manage these units in standard CoT makes this challenging.
This highlights the need for more fine-grained approaches to improve reasoning efficiency.

Building on this insight, we introduce \textbf{\method (\abbr)} to decode the ``think-then-answer'' CoT reasoning into a sequence of multi-turn interactions to enable the explicit control of the number of thinking units, where each turn contains a single thinking unit and an answer generated based on both the current and all preceding units.
Refer to \Cref{fig:intro} for an illustration of the paradigm shift. %
To implement \abbr, we adopt a pipeline combining SFT and GRPO.
We first convert conventional CoT traces into structured, multi-turn formats using GPT-4o~\citep{openai2024gpt4technicalreport} and then fine-tune the target model on such data. %
To further enhance efficiency, we apply GRPO to encourage the model to generate accurate responses within fewer reasoning turns, thereby reducing latency and computational costs.

To evaluate the effectiveness of \abbr, we conduct extensive experiments across a range of reasoning benchmarks. On DeepSeek-R1-Distill-Qwen-1.5B, \abbr reduces token usage by up to $\sim70\%$ and accelerates time to first token (TTFT) by $4.2\times$ on MATH-500, while maintaining over 95\% accuracy. Furthermore, \abbr demonstrates strong out-of-distribution generalization on this model, with token reductions of 69\% on AIME24 and 53\% on GPQA-Diamond. These results highlight the efficiency and broad applicability of \abbr in diverse reasoning scenarios.

\section{Related Work}
\label{sec:related}

\paragraph{Efficient Reasoning Paradigms}
The evolution of reasoning frameworks for LLMs has progressed significantly since the introduction of CoT prompting~\citep{wei2022chainofthought}. CoT has proven effective in enhancing LLMs' reasoning abilities by explicitly guiding models through intermediate reasoning steps~\citep{guo2025deepseek}, but this approach often leads to excessively lengthy outputs, resulting in high token consumption and increased latency~\citep{chiang2024overreasoningredundantcalculationlarge}. These inefficiencies have motivated researchers to explore more compact and efficient reasoning paradigms. One prominent line of work aims to reduce intermediate token usage without sacrificing reasoning quality. For example, methods like token skipping~\citep{xia2024tokenskip} and length-harmonizing pruning~\citep{luo2025o1prunerlengthharmonizingfinetuningo1like} have demonstrated significant reductions in token counts while maintaining strong task performance~\citep{fu2025reasoning}. These approaches directly target the redundancy challenge by refining the granularity of reasoning traces, thereby reducing overall token overhead. Another approach seeks to decouple the reasoning process from explicit token generation by leveraging continuous latent spaces. For instance, Token-Assorted Mixing~\citep{su2025tokenassortedmixinglatent} and Hidden Thinking frameworks~\citep{shen2025efficientreasoninghiddenthinking} aim to perform internal computations without generating extensive token sequences, achieving 3-5× faster processing speeds compared to conventional CoT~\citep{hao2025training}. This direction effectively compresses intermediate steps into compact latent representations, significantly improving efficiency. Additionally, several studies have explored integrating reasoning and non-reasoning models to enhance efficiency. For example, the C3OT system~\citep{kang2025c3ot} employs a multi-stage verification pipeline to reduce token redundancy, while speculative reasoning approaches~\citep{pan2025specreason} dynamically adjust the reasoning depth based on task complexity, further reducing token usage. Hybrid architectures like Hawkeye~\citep{she2025hawkeyeefficientreasoningmodelcollaboration} also leverage speculative decoding~\citep{zhang-etal-2024-draft} to balance accuracy and computational efficiency.

\paragraph{Reinforcement Learning for Reasoning Optimization}
Reinforcement learning (RL) has become an essential tool for optimizing LLM reasoning, providing precise control over decision-making processes. Group Relative Policy Optimization (GRPO)~\citep{shao2024deepseekmath} is one of the most influential methods in this domain, aligning reward signals with step-wise reasoning validity rather than simply final answer correctness. This strategy allows models to prioritize accurate intermediate steps, enhancing both response precision and computational efficiency. Building on this foundation, frameworks like DAPO~\citep{yu2025dapoopensourcellmreinforcement} and R1-Zero~\citep{liu2025understandingr1zeroliketrainingcritical} incorporate dynamic reward shaping and entropy-controlled exploration to further refine model outputs. These methods extend GRPO by introducing adaptive mechanisms that reduce token redundancy while maintaining high accuracy, making them particularly effective for complex reasoning tasks. Recent advancements have also focused on integrating search-based techniques to enhance reasoning efficiency. For instance, Search-R1~\citep{jin2025searchr1trainingllmsreason} combines Monte Carlo Tree Search with policy gradients to optimize reasoning path selection, reducing unnecessary token usage. Similarly, length-aware control frameworks like L1-Controller~\citep{aggarwal2025l1controllinglongreasoning} balance correctness and token efficiency through dual reward signals, achieving substantial latency reductions. Other approaches, such as R1-Searcher~\citep{song2025r1searcher}, incorporate dynamic halting mechanisms to automatically terminate unproductive reasoning chains, significantly improving efficiency in open-domain tasks. ThinkPrune~\citep{hou2025thinkprunepruninglongchainofthought} adopts length clipping to the reward function, pruning outputs to reduce redundancy.

\paragraph{Training-Based Efficiency Enhancements}
Training strategies have also played a critical role in improving reasoning efficiency. Supervised fine-tuning (SFT) methods like Thinking-Optimal Scaling~\citep{yang2025thinkingoptimalscalingtesttimecompute} align models with optimal solution trajectories, reducing token redundancy without compromising accuracy. This approach effectively reshapes the internal reasoning paths of models, ensuring more concise outputs. Hybrid training regimes have also gained traction, combining imitation learning and reinforcement learning to refine reasoning efficiency. For example, the SpecReason framework~\citep{pan2025specreason} employs a two-stage process, beginning with teacher-student distillation for foundational policy approximation, followed by adversarial reward shaping for fine-grained optimization. This blend of supervised and reinforcement learning techniques has proven effective in reducing token counts while maintaining response quality.

\section{Method}
\label{sec:method}

In this section, %
we first introduce the standard Chain-of-Thought (CoT) reasoning of Large reasoning models (LRMs) and briefly review Group Relative Policy Optimization (GRPO)~\cite{deepseekai2025deepseekr1incentivizingreasoningcapability}.
We then present an empirical study showing how redundant reasoning steps commonly arise in LRMs.
Finally, we outline \abbr, which reformulates the standard CoT into a multi-turn structure, and discuss how to leverage GRPO to encourage concise and effective multi-turn reasoning.

\begin{figure}[t]
    \centering
    \includegraphics[width=\linewidth]{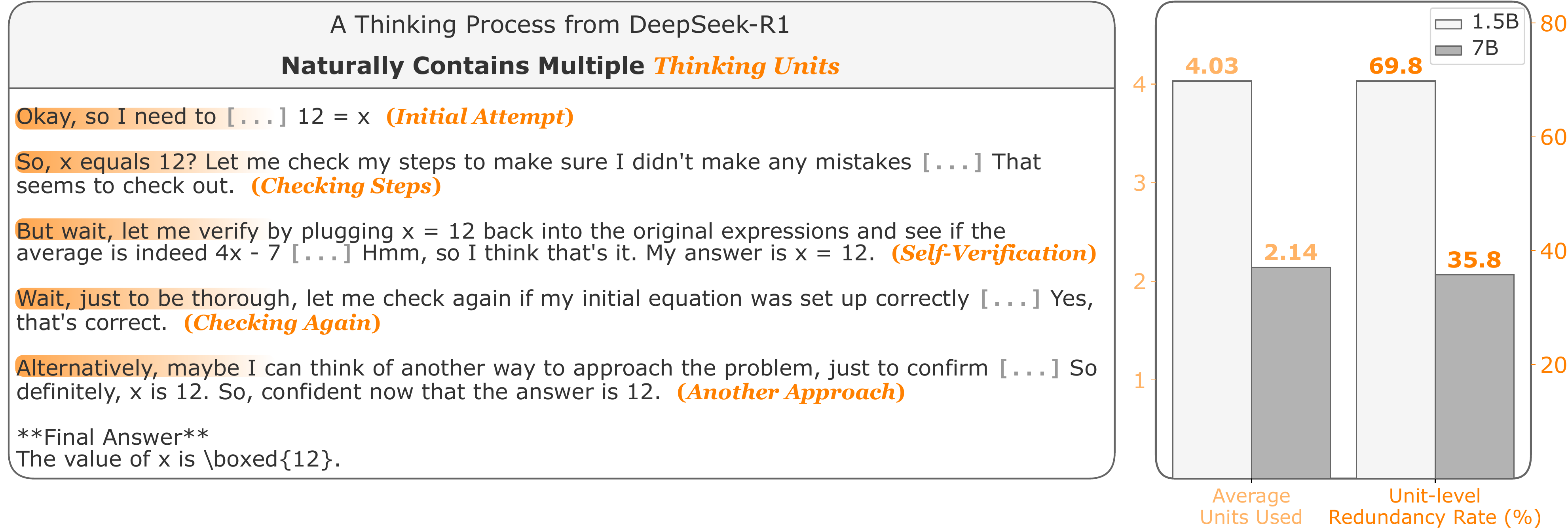}
    \caption{\textbf{Left:} An example of a standard CoT from DeepSeek-R1, naturally containing multiple discrete thinking units (the start of each new unit is marked with an orange highlight).
    \textbf{Right:} Empirical analysis of unit-level redundancy, which is calculated based on \Cref{eq:redundancy}, in R1-distilled models on the MATH-500 dataset, showing an average redundancy rate of 69.8\% for the 1.5B model and 35.8\% for the 7B model.
    }
    \label{fig:method-1}
\end{figure}

\subsection{Preliminary}
\label{sec:method-pre}

\paragraph{CoT for LRMs}
LRMs commonly adopt a ``think-then-answer'' paradigm for complex problem solving. Given a query $q$, an LRM typically produces an output $o$ of the form:
\begin{equation}
    q \rightarrow o = \verb|<think>|~t~\verb|</think>|~a~,
    \label{eq:cot}
\end{equation}
where $t$ denotes the internal thinking process, delimited by \verb|<think>| and \verb|</think>|, and $a$ is the final answer. The thinking process $t$ can be viewed as an exploration of the solution space and is naturally decomposed into multiple \emph{thinking units}---self-contained logical steps that can induce a candidate answer to $q$, with an example from DeepSeek-R1~\citep{guo2025deepseek} depicted in \Cref{fig:method-1} (left).
Formally, letting $u_i$ denote a thinking unit, there is $t = (u_1, u_2, \dots, u_n)$. %
These units may arise from (1) an initial attempt to solve the problem, (2) depth-wise exploration such as validation, backtracking, or correction along a single line of reasoning, or (3) breadth-wise search involving alternative methods or perspectives. Each unit can thus be interpreted as a path in the reasoning space, potentially building on previous steps, and may terminate with a provisional answer to the query.

However, current LRMs tend to employ numerous thinking units before gaining the final answer to solve the problem as `perfectly' as possible, causing significant inefficiency issues.

\paragraph{GRPO}
Let $\pi_\theta$ denote the current policy and $\pi_{\theta_{\mathrm{old}}}$ the reference policy from the previous iteration.
Given a query $q$, GRPO samples $G$ completions $o_1, \ldots, o_G$ and optimizes the objective:
\begin{equation}
\label{eq:grpo}
\mathbb{E}_{q,\, \{o_i\}_{i=1}^{G}} \left[
    \frac{1}{G} \sum_{i=1}^G {\color[HTML]{FF8000}\frac{1}{|o_i|}} \sum_{j=1}^{|o_i|}
        \min\left( \rho_{i,j} A_i,\ \operatorname{clip}(\rho_{i,j}, 1-\epsilon, 1+\epsilon) A_i \right)
\right],
\end{equation}
where $\rho_{i,j} = \frac{\pi_\theta(o_{i,j} \mid q, o_{i,<j})}{\pi_{\theta_{\mathrm{old}}}(o_{i,j} \mid q, o_{i,<j})}$ is the ratio between the new and old policies for token $j$ in sequence $o_i$ and $|o_i|$ is the sequence length.
$A_i$ is the group-standardized advantage:
\begin{equation}
A_i = \frac{R(o_i) - \mathrm{mean}(\{R(o_1), \ldots, R(o_G)\})}{\mathrm{std}(\{R(o_1), \ldots, R(o_G)\})}~,
\label{eq:advantage}
\end{equation}
where $R$ denotes the reward function, and $\mathrm{mean}(\{r_1, \ldots, r_G\})$ and $\mathrm{std}(\{r_1, \ldots, r_G\})$ represent the mean and standard deviation of group rewards, respectively. For clarity, we omit the KL regularization term, as it is not the focus of our analysis.

\subsection{Unit-Level Redundancy in LRMs}
\label{sec:method-redundancy}

Before devoting to reducing the number of thinking units of LRMs, we first systematically investigate the \emph{unit-level redundancy}, which is intuitively high considering the repeated depth-wise validations or breadth-wise explorations of alternative solution paths, even after repeatedly arriving at essentially the same valid answer,  in long CoTs.

Concretley, we conducted a detailed analysis using DeepSeek-R1-Distill-Qwen-1.5B/7B~\citep{deepseekai2025deepseekr1incentivizingreasoningcapability}. We extracted their CoT traces from the MATH~\citep{lightman2023lets} and GSM8K~\citep{cobbe2021gsm8k} training sets (restricted to correctly answered examples), and segmented each trace into discrete thinking units using GPT-4o~\citep{openai2024gpt4technicalreport} (see \Cref{sec:prompt} for details).

For each segmented trace $t = (u_1, u_2, \ldots, u_n)$, we constructed prefix sub-traces $t_{\leq k} = (u_1, \ldots, u_k)$ for $1 \leq k \leq n$. We then prompted the model to generate an intermediate answer $a_k$ by appending a special stop token \verb|</think>| after $t_{\leq k}$ %
given the current partial reasoning:
\begin{equation}
    q \rightarrow o_k = \verb|<think>|~t_{\leq k}~\verb|</think>|~a_k~,\quad k=1,\cdots,n~.
\end{equation}

To quantify unit-level redundancy, we define the minimal sufficient prefix $t_{\leq n^*}$ as the shortest prefix that leads to a correct final answer. The \emph{unit-level redundancy rate} is then defined as:
\begin{equation}
    \text{URR} = \frac{n - n^*}{n} \cdot \mathbbm{1}_{a_n \text{ is correct}}~,
    \label{eq:redundancy}
\end{equation}
where $n$ is the total number of thinking units and $n^*$ is the minimal number required for correctness.
A higher URR indicates a greater proportion of unnecessary reasoning steps.

Our empirical results, summarized in \Cref{fig:method-1} (right), show that the average unit-level redundancy rates are $69.8\%$ for the 1.5B model and $35.8\%$ for the 7B model.
This reveals that a significant portion of the reasoning process in current LRMs is redundant for solving the problem, underscoring the potential for substantial efficiency gains by explicitly mitigating unit-level redundancy.

\begin{figure}[t]
    \centering
    \includegraphics[width=\linewidth]{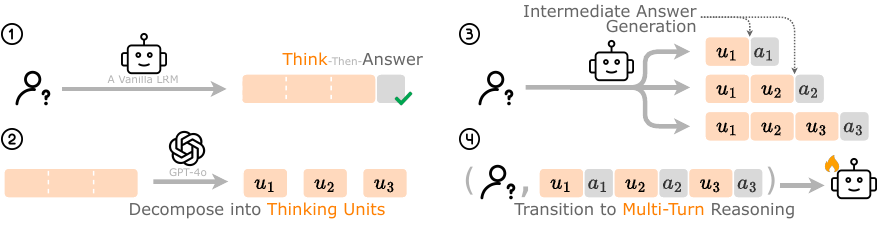}
    \caption{
Transforming think-then-answer LRMs into a multi-turn reasoning paradigm, consisting of four steps:
(1) Rejection sampling to filter out responses with correct final answers;
(2) Unit segmentation using GPT-4o to divide CoTs into discrete reasoning units;
(3) Intermediate answer completion to extract answers ($a_k$) for each prefix sub-trace ($t_{\le k}$);
and (4) SFT to align LRMs with the multi-turn format.
}
    \label{fig:method-2}
\end{figure}

\subsection{\method (\abbr)}
\label{sec:method-mind}

Our basic notion is that the model should not be that cautious.
Given that ``done is better than perfect'', we aim to let the model yield a candidate answer as soon as possible.
Besides, we would also like to penalize the unit-level redundancy.
\abbr realizes these through two key innovations.

\paragraph{Multi-Turn CoT Reformulation}
\abbr first employs supervised fine-tuning (SFT) to shift the reasoning paradigm from ``think-then-answer'' (i.e., \Cref{eq:cot}) to a structured multi-turn format:
\begin{equation}
    \verb|<think>|~u_1~\verb|</think>|~a_1\
    \verb|<think>|~u_2~\verb|</think>|~a_2\
    \cdots
    \verb|<think>|~u_n~\verb|</think>|~a_n\ ,
    \label{eq:multi-turn}
\end{equation}
where the thinking units $(u_1, u_2, \ldots, u_n)$ in the original CoT $t$ are distributed into a sequence of \emph{reasoning turns}.
Each turn also includes an intermediate answer $a_k$. %

To construct the training data for multi-turn SFT, we first segment the original thinking process $t$ into $(u_1, u_2, \ldots, u_n)$, and then generate an intermediate answer $a_k$ after each $u_k$, as described in \Cref{sec:method-redundancy}. %
The overall pipeline is illustrated in \Cref{fig:method-2}.

After training, the learned multi-turn LRM enables flexible management of the thinking units (e.g., choose to continue or abort from the reasoning by manipulating the token $\verb|</think>|$), but we empirically observe that when applying no control, the model tends to generate even more output tokens than the original one (see \Cref{tab:compare-sft-mind}).
This is because SFT primarily reshapes the reasoning format without directly addressing unit-level redundancy, and $a_k$ incurs further token usage.
To bridge the gap, we suggest leveraging GRPO to prioritize efficient reasoning traces. %

\paragraph{Reducing Reasoning Turns via GRPO}
We define a reward function $R$ comprises three components for GRPO:
\begin{equation}
R = \mathcal{R}_{\text{format}} + \mathcal{R}_{\text{accuracy}} + \mathcal{R}_{\text{unit}}~.
\end{equation}
In detail, they are:
\hlgray{(1) Format Consistency Reward $\mathcal{R}_{\text{format}}$}, which ensures that the generated output adheres to the multi-turn structure described in \Cref{eq:multi-turn}.
\hlgray{(2) Answer Accuracy Reward $\mathcal{R}_{\text{accuracy}}$}, which rewards the model for producing a correct final answer, as determined by matching $a_n$ to the ground truth.
\hlgray{(3) Unit Compactness Reward $\mathcal{R}_{\text{unit}}$}, which penalizes cases where a single reasoning unit contains multiple exploratory trajectories and thus encourages a clear separation between reasoning turns.
See ~\Cref{sec:exp-ablation} for further analysis of this component.
The specific weights for each reward component are detailed in \Cref{tab:exp-setup-1}.

Note that we do not introduce an explicit reward term regarding the number of turns, because
GRPO inherently introduces an implicit bias toward generating shorter CoTs that yield correct answers.
As shown in \Cref{eq:grpo}, for a fixed advantage $A_i$, the per-token normalization $\sfrac{1}{|o_i|}$ results in larger per-token updates for shorter outputs~\citep{lin2025cppo,yu2025dapoopensourcellmreinforcement,liu2025understandingr1zeroliketrainingcritical}, thereby encouraging the model to produce more concise and efficient completions.
This effect is particularly pronounced in LRMs, which typically possess strong reasoning capabilities and can generate multiple correct yet diverse completions per group during training.
Thus, the GRPO framework naturally incentivizes the model to favor responses with fewer reasoning turns. This behavior is empirically validated in \Cref{fig:countofturn}, where we observe a substantial reduction in the number of reasoning turns following GRPO training.

\section{Experiments}
\label{sec:exp}

In this section, we evaluate the efficiency of \abbr across several benchmarks. \Cref{sec:exp-setup} describes the experimental setup. \Cref{sec:exp-main} presents the main results, focusing on token reduction, accuracy, and latency. Ablation studies and additional discussion are provided in \Cref{sec:exp-ablation}.

\subsection{Setup}
\label{sec:exp-setup}

\begin{table}[h]
\centering
\begin{minipage}{0.55\linewidth}
\centering
\caption{Reward function value settings.}
\label{tab:exp-setup-1}
\begin{tabular}{lccc}
\toprule
&  $\mathcal{R}_{\text{format}}$  &  $\mathcal{R}_{\text{accuracy}}$  &  $\mathcal{R}_{\text{unit}}$  \\
\midrule
Compliance & +1 & +2 & 0 \\
Non-Compliance & -1 & -2 & -0.3 \\
\bottomrule
\end{tabular}
\end{minipage}%
\begin{minipage}{0.4\linewidth}
\centering
\setlength{\tabcolsep}{10pt}
\caption{Training data sizes.}
\label{tab:exp-setup-2}
\begin{tabular}{lcc}
\toprule
& 1.5B & 7B \\
\midrule
SFT & 3610 & 3532 \\
GRPO & 7500 & 7500 \\
\bottomrule
\end{tabular}
\end{minipage}
\end{table}

\paragraph{Training Details}
The training process for \abbr consists of two key phases, as described in \Cref{sec:method-mind}. The first SFT phase is conducted using the LLaMA-Factory repository~\citep{zheng2024llamafactory}. We perform full-parameter fine-tuning for 2 epochs with a learning rate of 5e-5. The second GRPO phase leverages the veRL repository~\citep{sheng2024hybridflow}. During this phase, we train for 1 epoch with an actor learning rate of 1e-6. For each training step, 10 roll-out completions are generated for each sample, with all other hyperparameters set to the default values provided by veRL. The reward function described in Section~\ref{sec:method-mind} is adopted with the weight configurations listed in Table~\ref{tab:exp-setup-1}.

\paragraph{Models \& Datasets}
We conduct our experiments using DeepSeek-R1-Distill-Qwen-1.5B/7B~\citep{deepseekai2025deepseekr1incentivizingreasoningcapability}. For SFT, the training data consists of questions from the GSM8K~\citep{cobbe2021gsm8k} and MATH~\citep{lightman2023lets} training sets. Model-generated responses are filtered via rejection sampling to retain only correct answers, then pre-processed as shown in \Cref{fig:method-2}. For GRPO, we use the MATH training set exclusively, with sample sizes detailed in \Cref{tab:exp-setup-2}.
We evaluate on both in-distribution (MATH-500~\citep{lightman2023lets}) and out-of-distribution benchmarks, including AIME24~\citep{aime2024}, AMC23~\citep{amc23}, and GPQA-Diamond~\citep{rein2023gpqa}, to assess generalization.

\paragraph{Baselines}
To assess the efficiency of our method, we compare against the following baselines:
\hlgray{Original LRM}: The base models used in this work, DeepSeek-R1-Distill-Qwen-1.5B and 7B.
\hlgray{ThinkPrune}~\citep{hou2025thinkprunepruninglongchainofthought}: Adds length clipping to the GRPO reward and is trained on the AIME-AMC subset, progressively pruning outputs at the token level to reduce response length.
\hlgray{DEER}~\citep{yang2025dynamicearlyexitreasoning}: A training-free approach that detects ``action transition points'' (e.g., ``Wait,'' ``Alternatively,'' ``Hmm'') to trigger answer generation, and halts decoding when the mean token probability surpasses a confidence threshold.
\hlgray{Dynasor}~\citep{fu2025reasoning}: Periodically inserts probes (e.g., every 32, 64, or 128 tokens) to extract intermediate answers and assess their consistency, enabling early termination of generation.

\begin{table}[t]
\centering
\caption{
Performance comparison of various baselines and our proposed method, \abbr, across four reasoning benchmarks: MATH-500, AIME24, AMC23, and GPQA-Diamond. The table reports both accuracy (Acc.; higher is better) and average output token usage (Tokens; lower is better) for each model. Results are shown for both 1.5B and 7B parameter configurations, covering the original LRM (DeepSeek-R1-Distill-Qwen-1.5B and 7B), ThinkPrune~\citep{hou2025thinkprunepruninglongchainofthought}, Dynasor~\citep{fu2025reasoning}, DEER~\citep{yang2025dynamicearlyexitreasoning}, and our method, \abbr. Note that for \abbr, GRPO is performed only on the MATH training set, making MATH-500 in-domain and the others out-of-domain.
As shown in the table, \abbr consistently achieves competitive or superior accuracy while significantly reducing token usage, demonstrating its effectiveness for efficient and generalizable reasoning.
}
\resizebox{\textwidth}{!}{ %
\begin{tabular}{l *{4}{>{\centering\arraybackslash}p{14mm} >{\centering\arraybackslash}p{15mm}}}
\toprule[1.5pt]
 & \multicolumn{2}{c}{MATH-500} & \multicolumn{2}{c}{AIME24} & \multicolumn{2}{c}{AMC23} & \multicolumn{2}{c}{GPQA-Diamond} \\
\cmidrule(lr){2-3} \cmidrule(lr){4-5} \cmidrule(lr){6-7} \cmidrule(lr){8-9}
 & Acc.$\uparrow$ & Tokens$\downarrow$ & Acc.$\uparrow$ & Tokens$\downarrow$ & Acc.$\uparrow$ & Tokens$\downarrow$ & Acc.$\uparrow$ & Tokens$\downarrow$ \\
\midrule[1.1pt]
\multicolumn{9}{c}{\textbf{1.5B}} \\
\midrule[0.6pt]
Original LRM & 85.4 & 5389 & 26.7 & 15177 & 67.5 & 9956 & 32.3 & 9842 \\
\addlinespace[5pt]
ThinkPrune~\citep{hou2025thinkprunepruninglongchainofthought} & 83.2\makebox[0pt][l]{\hspace{0.2em}\scriptsize\textsubscript{-2.6\%}} & 1938\makebox[0pt][l]{\hspace{0.2em}\scriptsize\textsubscript{-64\%}} & 27.1\makebox[0pt][l]{\hspace{0.2em}\scriptsize\textsubscript{+1.5\%}} & 5631\makebox[0pt][l]{\hspace{0.2em}\scriptsize\textsubscript{-63\%}} & 73.2\makebox[0pt][l]{\hspace{0.2em}\scriptsize\textsubscript{+8.4\%}} & 3039\makebox[0pt][l]{\hspace{0.2em}\scriptsize\textsubscript{-70\%}} & - & - \\
DEER~\citep{yang2025dynamicearlyexitreasoning} & 73.2\makebox[0pt][l]{\hspace{0.2em}\scriptsize\textsubscript{-14.3\%}} & 1118\makebox[0pt][l]{\hspace{0.2em}\scriptsize\textsubscript{-79\%}} & 20.0\makebox[0pt][l]{\hspace{0.2em}\scriptsize\textsubscript{-25.1\%}} & 3302\makebox[0pt][l]{\hspace{0.2em}\scriptsize\textsubscript{-78\%}} & 47.5\makebox[0pt][l]{\hspace{0.2em}\scriptsize\textsubscript{-29.6\%}} & 2384\makebox[0pt][l]{\hspace{0.2em}\scriptsize\textsubscript{-76\%}} & 5.6\makebox[0pt][l]{\hspace{0.2em}\scriptsize\textsubscript{-82.7\%}} & 4128\makebox[0pt][l]{\hspace{0.2em}\scriptsize\textsubscript{-58\%}} \\
\addlinespace[3pt]
\rowcolor{gray!15}
\abbr & 82.8\makebox[0pt][l]{\hspace{0.2em}\scriptsize\textsubscript{-3.0\%}} & 1719\makebox[0pt][l]{\hspace{0.2em}\scriptsize\textsubscript{-68\%}}
 & 30.0\makebox[0pt][l]{\hspace{0.2em}\scriptsize\textsubscript{+12.4\%}} & 4856\makebox[0pt][l]{\hspace{0.2em}\scriptsize\textsubscript{-68\%}} & 77.5\makebox[0pt][l]{\hspace{0.2em}\scriptsize\textsubscript{+14.8\%}} & 2384\makebox[0pt][l]{\hspace{0.2em}\scriptsize\textsubscript{-76\%}} & 31.3\makebox[0pt][l]{\hspace{0.2em}\scriptsize\textsubscript{-3.1\%}} & 4690\makebox[0pt][l]{\hspace{0.2em}\scriptsize\textsubscript{-52\%}} \\
\midrule[1.1pt]
\multicolumn{9}{c}{\textbf{7B}} \\
\midrule[0.6pt]
Original LRM & 93.0 & 3928 & 50.0 & 14107 & 90.0 & 6076 & 50.5 & 8390 \\
\addlinespace[5pt]
Dynasor~\citep{fu2025reasoning} & 88.5\makebox[0pt][l]{\hspace{0.2em}\scriptsize\textsubscript{-4.8\%}} & 2591\makebox[0pt][l]{\hspace{0.2em}\scriptsize\textsubscript{-34\%}} & 47.7\makebox[0pt][l]{\hspace{0.2em}\scriptsize\textsubscript{-4.6\%}} & 8760\makebox[0pt][l]{\hspace{0.2em}\scriptsize\textsubscript{-38\%}} & 87.1\makebox[0pt][l]{\hspace{0.2em}\scriptsize\textsubscript{-3.2\%}} & 4913\makebox[0pt][l]{\hspace{0.2em}\scriptsize\textsubscript{-19\%}} & - & - \\
DEER~\citep{yang2025dynamicearlyexitreasoning} & 87.4\makebox[0pt][l]{\hspace{0.2em}\scriptsize\textsubscript{-6.0\%}} & 975\makebox[0pt][l]{\hspace{0.2em}\scriptsize\textsubscript{-75\%}} & 33.3\makebox[0pt][l]{\hspace{0.2em}\scriptsize\textsubscript{-33.4\%}} & 3235\makebox[0pt][l]{\hspace{0.2em}\scriptsize\textsubscript{-77\%}} & 82.5\makebox[0pt][l]{\hspace{0.2em}\scriptsize\textsubscript{-8.3\%}} & 1622\makebox[0pt][l]{\hspace{0.2em}\scriptsize\textsubscript{-73\%}} & 27.3\makebox[0pt][l]{\hspace{0.2em}\scriptsize\textsubscript{-45.9\%}} & 2265\makebox[0pt][l]{\hspace{0.2em}\scriptsize\textsubscript{-73\%}} \\
\addlinespace[3pt]
\rowcolor{gray!15}
\abbr & 91.6\makebox[0pt][l]{\hspace{0.2em}\scriptsize\textsubscript{-1.5\%}} & 2859\makebox[0pt][l]{\hspace{0.2em}\scriptsize\textsubscript{-27\%}} & 46.7\makebox[0pt][l]{\hspace{0.2em}\scriptsize\textsubscript{-6.6\%}} & 7258\makebox[0pt][l]{\hspace{0.2em}\scriptsize\textsubscript{-49\%}} & 95.0\makebox[0pt][l]{\hspace{0.2em}\scriptsize\textsubscript{+5.6\%}} & 3777\makebox[0pt][l]{\hspace{0.2em}\scriptsize\textsubscript{-38\%}} & 53.0\makebox[0pt][l]{\hspace{0.2em}\scriptsize\textsubscript{+5.0\%}} & 6845\makebox[0pt][l]{\hspace{0.2em}\scriptsize\textsubscript{-18\%}} \\
\bottomrule[1.5pt]
\end{tabular}
}
\label{tab:benchmark-results}
\end{table}

\paragraph{Evaluation Metrics}
We evaluate \abbr using three primary metrics: accuracy, average output token usage, and time-to-first-token (TTFT). TTFT measures the time it takes for the model to generate the first answer token of the response, from when the prompt was sent---a key determinant of user experience. The evaluations are conducted using the Open-R1 evaluation scripts~\citep{openr1}, with a maximum sequence length of 32,768 tokens, a temperature setting of 0.6, and a top-p value of 0.95, running on four NVIDIA A100 GPUs.

\subsection{Main Results}
\label{sec:exp-main}

\paragraph{Reducing Output Tokens for Efficient Reasoning}
After training the 1.5B and 7B multi-turn reasoning models as described in \Cref{sec:exp-setup}, we evaluated their token efficiency across a range of reasoning benchmarks. The results, summarized in \Cref{tab:benchmark-results}, show that \abbr consistently reduces output token usage while maintaining strong performance.
On in-domain MATH-500, \abbr lowers the average token usage to 1719 for the 1.5B model—a 68\% reduction from the Original LRM (5389 tokens)—while achieving 82.8\% accuracy. Although ThinkPrune attains similar accuracy (83.2\%), it requires more tokens (1938). DEER achieves the lowest token usage (1118), but with a substantial accuracy drop to 73.2\%. For the 7B model, \abbr reduces average token usage by 27\% compared to the Original LRM (2859 vs. 3928), with a high accuracy of 91.6\%, outperforming both Dynasor and DEER in the balance of accuracy and efficiency.
\abbr's efficiency generalizes well to out-of-domain benchmarks. For example, on AMC23 (1.5B), \abbr reaches 77.5\% accuracy with 2384 tokens, substantially outperforming ThinkPrune and DEER in both accuracy and token reduction. Similar trends are observed on AIME24 and GPQA-Diamond.
These results demonstrate that \abbr effectively eliminates unnecessary reasoning steps, producing concise, efficient outputs without compromising performance.

\begin{figure}[t]
    \centering
    \begin{minipage}[t]{0.34\textwidth}
        \centering
        \includegraphics[width=\linewidth]{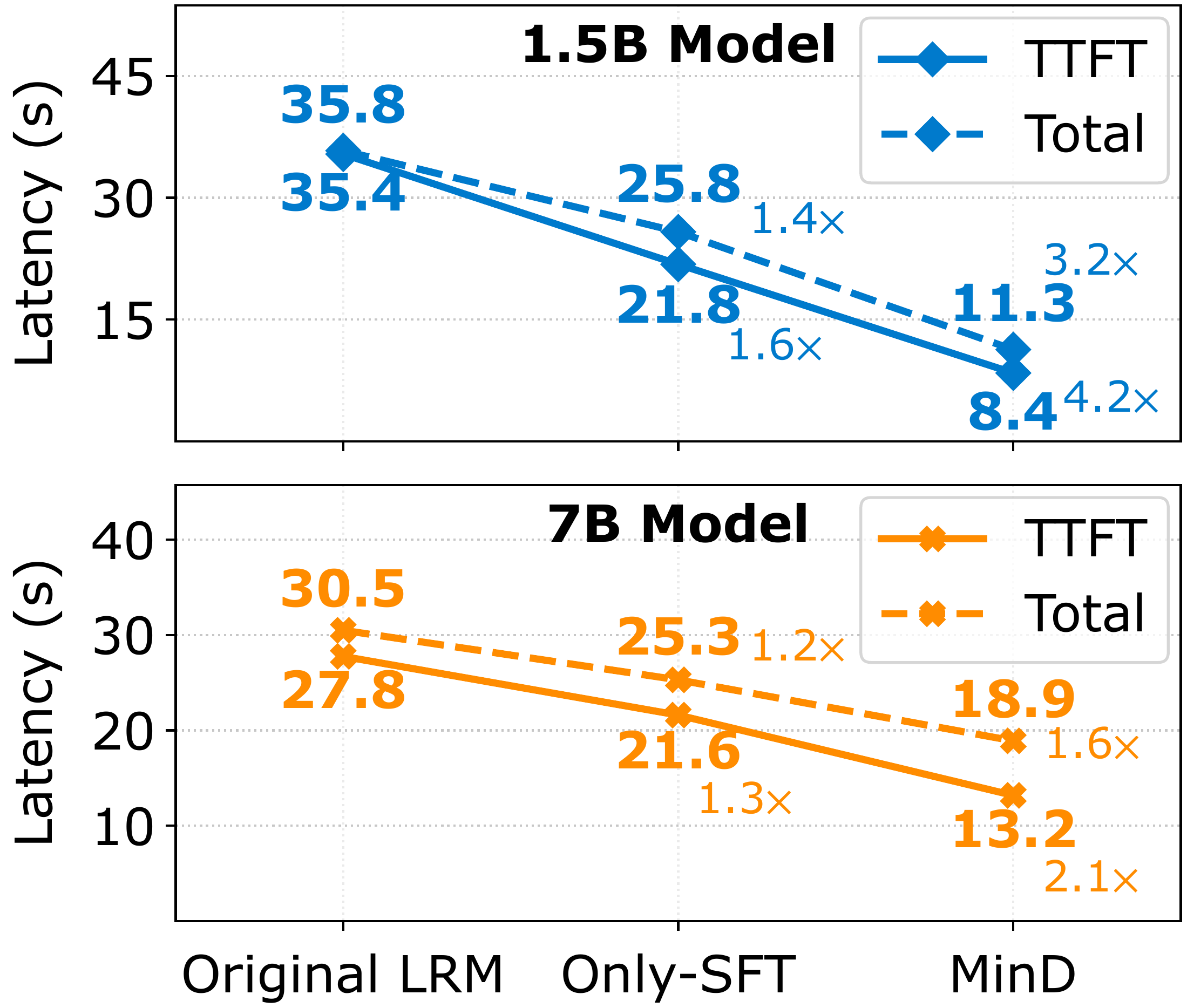}
        \caption{TTFT (time to first token) and total latency of two DeepSeek-R1-distilled models on MATH-500. \abbr achieves up to 4.2$\times$ (1.5B) and 2.1$\times$ (7B) speedups over the original LRMs in TTFT, and 3.2$\times$ (1.5B) and 1.6$\times$ (7B) in total latency.}
        \label{fig:first-token}
    \end{minipage}\hspace{0.04\textwidth}
    \begin{minipage}[t]{0.6\textwidth}
        \centering
        \includegraphics[width=\linewidth]{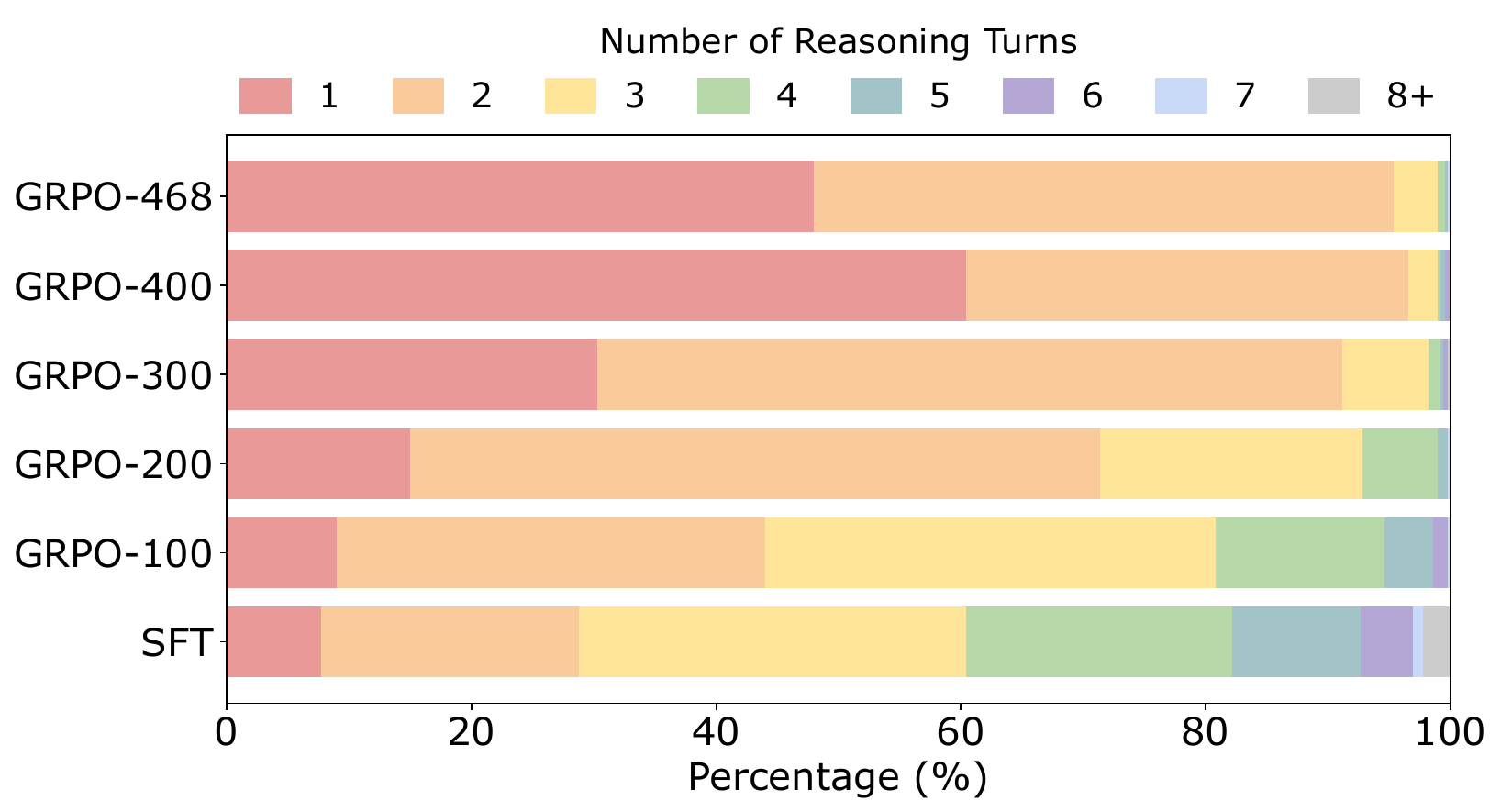}
        \caption{
        The distribution of reasoning turns for \abbr at different training stages (1.5B model) on the MATH-500 dataset. Each bar represents a model checkpoint, including the SFT model and successive GRPO training steps. As GRPO training progresses, the number of reasoning turns per output decreases and becomes increasingly concentrated at 1 or 2 turns (highlighted in red and orange), demonstrating the effectiveness of GRPO in mitigating reasoning redundancy.
        }
        \label{fig:countofturn}
    \end{minipage}
\end{figure}

\paragraph{Reducing TTFT and Total Latency}
The TTFT and total response latency for the original R1-distilled LRMs and our \abbr models are summarized in \Cref{fig:first-token}. As shown, \abbr significantly reduces both TTFT and total latency across both model sizes.
For the 1.5B configuration, the original 1.5B model requires 35.4s TTFT, which drops to 21.8s after SFT and further to 8.4s with \abbr, resulting in a 4.2$\times$ speedup. The total latency is similarly reduced from 35.8s (original) to 25.8s (SFT) and 11.3s (\abbr), a 2.1$\times$ improvement.
For the 7B model, TTFT decreases from 27.8s (original) to 21.6s (SFT) and 13.2s (\abbr), achieving a 2.1$\times$ speedup. The total latency is reduced from 30.5s to 25.3s and 18.9s, for a 1.6$\times$ speedup.
These results show that \abbr shortens both the time to first answer token and the overall response latency, making the models more responsive.

\begin{table}[t]
\centering
\caption{
Comparison of different training strategies on DeepSeek-R1-Distill-Qwen-1.5B.
Original LRM refers to the pretrained baseline.
SFT-Only applies only the supervised fine-tuning step from MinD.
Non-Multi-Turn applies GRPO without explicit multi-turn segmentation.
MinD denotes our full method with both multi-turn segmentation and GRPO.
Acc.$\uparrow$ indicates accuracy (higher is better), and Tokens$\downarrow$ indicates average output length (lower is better).
}
\label{tab:compare-sft-mind}
\resizebox{\textwidth}{!}{ %
\begin{tabular}{lcccccccc}
\toprule
& \multicolumn{2}{c}{Original LRM}
& \multicolumn{2}{c}{SFT-Only}
& \multicolumn{2}{c}{Non-Multi-Turn}
& \multicolumn{2}{c}{MinD} \\
\cmidrule(lr){2-3} \cmidrule(lr){4-5} \cmidrule(lr){6-7} \cmidrule(lr){8-9}
& Acc.$\uparrow$ & Tokens$\downarrow$ & Acc.$\uparrow$ & Tokens$\downarrow$ & Acc.$\uparrow$ & Tokens$\downarrow$ & Acc.$\uparrow$ & Tokens$\downarrow$ \\
\midrule
MATH-500 & 85.4 & 5389 & 82.8 & 5655 & 82.0 & 1866 & 82.8 & 1719 \\
AIME24   & 26.7 & 15177 & 26.7 & 20675 & 20.0 & 7654 & 30.0 & 4856 \\
AMC23    & 67.5 & 9956  & 77.5 & 8409  & 65.0 & 3415 & 77.5 & 2384 \\
GPQA-Diamond     & 32.3 & 9842  & 28.3 & 12501 & 28.8 & 3397 & 37.4 & 4345 \\

\bottomrule
\end{tabular}
}
\end{table}

\subsection{Discussion \& Ablation}
\label{sec:exp-ablation}

\paragraph{GRPO is Crucial for Efficient Reasoning}
As discussed in \Cref{sec:method-mind}, SFT alone does not guarantee efficient reasoning. To demonstrate this, we compare the performance of models after SFT and after the full \abbr pipeline, as shown in \Cref{tab:compare-sft-mind}. The results reveal that SFT-only training often increases average output token usage relative to the original LRM. In contrast, applying GRPO further leads to substantial reductions in token usage while preserving accuracy, underscoring the essential role of GRPO in enabling concise and effective reasoning.

\begin{figure}[ht]
    \centering
    \includegraphics[width=\linewidth]{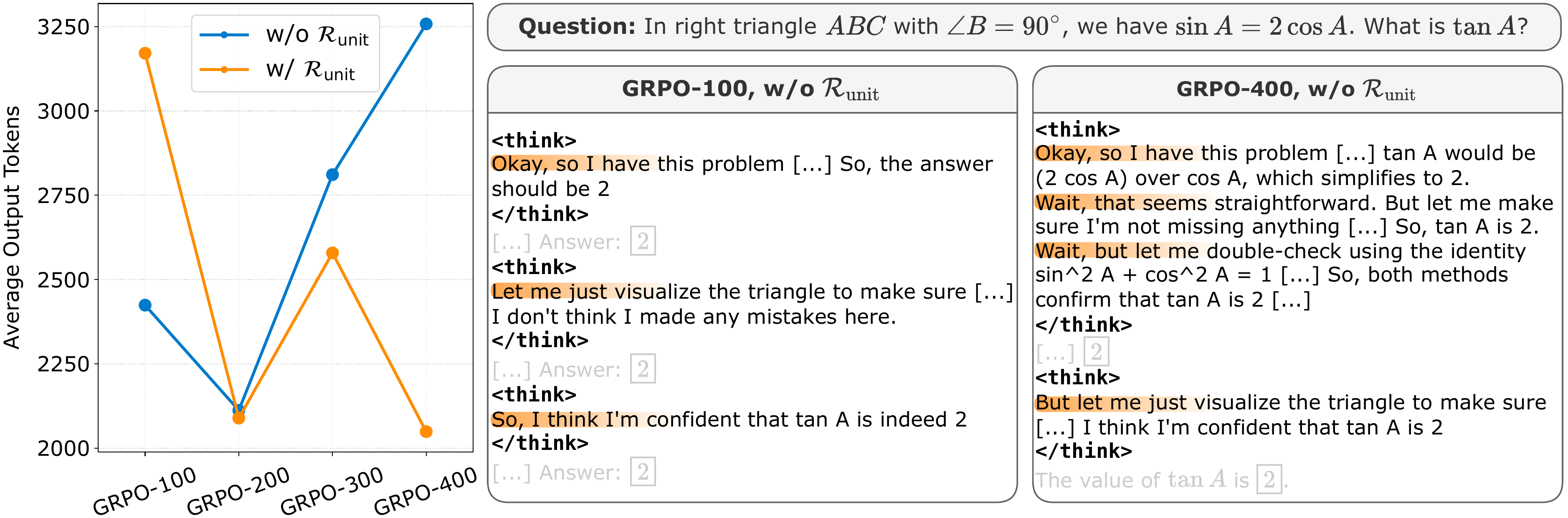}
    \caption{\textbf{Left:} Comparison of GRPO training with and without $\mathcal{R}_{\text{unit}}$ on MATH-500 for different 1.5B model checkpoints, showing Average Output Tokens for each. Removing $\mathcal{R}_{\text{unit}}$ leads to instability and collapse in output length.
\textbf{Right:} An illustrative case comparing the outputs of GRPO-100-step and GRPO-400-step checkpoints trained without $\mathcal{R}_{\text{unit}}$. While the earlier checkpoint (GRPO-100) maintains clear multi-turn reasoning, the later checkpoint (GRPO-400) exhibits several thinking units within a single turn (the start of each new unit is marked with an orange highlight), demonstrating that omitting $\mathcal{R}_{\text{unit}}$ results in blurred step boundaries and loss of controllable, structured reasoning.}
    \label{fig:r-unit}
\end{figure}

\paragraph{Role of $\mathcal{R}_{\text{unit}}$ in Maintaining Multi-Turn Reasoning}
As discussed in \Cref{sec:method-mind} and detailed in \Cref{tab:exp-setup-1}, our GRPO framework introduces a Unit Compactness Reward, $\mathcal{R}_{\text{unit}}$, to enforce that each reasoning turn contains only a single, coherent exploratory trajectory. This mechanism is essential for preventing the model from degenerating into the original monolithic think-then-answer style---a common outcome under GRPO's token-level averaging (\Cref{sec:method-mind}), which tends to favor shorter correct outputs. Without a specific penalty for multi-trajectory turns, the model may skip intermediate answers, collapsing the multi-turn reasoning structure into a single-block CoT.
To counteract this, $\mathcal{R}_{\text{unit}}$ penalizes reasoning turns that contain multiple exploratory trajectories, detected by linguistic cues such as phrases like ``double-check.'' This strategy encourages each turn to contain only one exploratory trajectory---especially in the critical first turn---without requiring external supervision, and thus maintains the multi-turn paradigm throughout training. The impact of $\mathcal{R}_{\text{unit}}$ is demonstrated in \Cref{fig:r-unit}, which shows how its absence leads to a collapse in output structure and length.

\paragraph{\abbr Effectively Alleviates Redundancy}
To demonstrate the effectiveness of GRPO in reducing redundancy, we plotted the distribution of reasoning turns for SFT and GRPO models on the MATH-500 dataset, as shown in \Cref{fig:countofturn}. The figure clearly illustrates that GRPO significantly reduces the number of reasoning turns, indicating a more compact and efficient reasoning process compared to the purely SFT-trained models.
Additionally, from the data in \Cref{tab:benchmark-results}, GRPO reduces the average output tokens on MATH-500 by 68.1\% for the 1.5B model and 27.2\% for the 7B model, compared to their respective original LRMs. This aligns well, though not directly, with the redundancy rates of 69.8\% and 35.8\% for these models, as reported in \Cref{fig:method-1} (Right). While these figures cannot be directly equated, they collectively indicate that \abbr, through GRPO, substantially alleviates redundancy, resulting in more concise and efficient outputs.

\paragraph{The Importance of Multi-Turn Structure}
To evaluate the impact of the multi-turn design, we performed SFT using responses from the original distilled-1.5B model, without applying any multi-turn segmentation (i.e., using the same question set as in step (1) of \Cref{fig:method-2}), followed by GRPO with only the format and outcome rewards. As shown in \Cref{tab:compare-sft-mind}, the Non-Multi-Turn model achieves comparable results to \abbr on in-distribution MATH-500, but exhibits a notable drop in accuracy and only marginal reductions in token usage on out-of-distribution benchmarks. We hypothesize that, under the conventional CoT format, models lack the flexibility to adjust the number of thinking units, making it difficult to learn a reasoning process that is both controllable and generalizable.

Additional discussion can be found in \Cref{sec:more results}.

\section{Conclusion}

In this paper, we introduced Multi-Turn Decomposition (MinD), an efficient method for improving the reasoning efficiency of large language models. By structuring the reasoning process into multi-turn steps, MinD significantly reduces token usage and response latency while maintaining strong performance across various reasoning tasks. Our results demonstrate that structured reasoning provides a practical solution to challenges such as slow response times and high computational costs in large language models.

\section{Limitation}
\label{sec:limitation}
Our work is limited by experiments on only 1.5B and 7B models and a primary focus on mathematical reasoning. Future directions include scaling to larger models, expanding to other reasoning domains, and developing adaptive multi-turn strategies that adjust the number of turns based on problem difficulty or user preference.

\bibliographystyle{plain}
\bibliography{refs}

\appendix
\crefalias{section}{appendix}
\crefalias{subsection}{subappendix}

\section{Word Frequency Analysis of Thinking Units}
\label{sec:more results}
In this section, we collect and compare the number of distinct words representing thinking units in DeepSeek-R1-Distill-1.5B, including the Original LRM, Non-Multi-Turn (GRPO applied without explicit multi-turn segmentation)
, and MinD. Although these words do not precisely correspond to the number of actual thinking units, they serve as a meaningful proxy and offer indicative insights into their distribution(see \Cref{tab:wordcount} for details).

\begin{table}[htbp]
\centering
\caption{The frequency of words representing thinking units in outputs generated by Original LRM, Non-Multi-Turn and MinD across MATH-500, AIME24 and AMC23.}
\label{tab:wordcount}
\begin{tabular}{lccccc}
\toprule[1.5pt]
~ & Wait & Alternatively & double-check & check & verify \\
\midrule[1.1pt]
\multicolumn{6}{c}{\textbf{MATH-500}} \\
\midrule[0.6pt]
Original LRM & 13993 & 2206 & 368 & 1272 & \textbf{124} \\
Non-Multi-Turn & 1822 & 333 & 41 & \textbf{347} & 193 \\
\rowcolor{gray!15}
MinD & \textbf{1651} & \textbf{237} & \textbf{10} & 434 & 249 \\
\midrule[1.1pt]
\multicolumn{6}{c}{\textbf{AIME24}} \\
\midrule[0.6pt]
Original LRM & 3742 & 415 & 20 & 215 & 17 \\
Non-Multi-Turn & 317 & 67 & \textbf{0} & 45 & 19 \\
\rowcolor{gray!15}
MinD & \textbf{211} & \textbf{45} & \textbf{0} & \textbf{34} & \textbf{8} \\
\midrule[1.1pt]
\multicolumn{6}{c}{\textbf{AMC23}} \\
\midrule[0.6pt]
Original LRM & 2302 & 385 & 35 & 205 & 45 \\
Non-Multi-Turn & 246 & 38 & 3 & \textbf{42} & \textbf{17} \\
\rowcolor{gray!15}
MinD & \textbf{215} & \textbf{30} & \textbf{0} & 50 & 22 \\
\bottomrule[1.5pt]
\end{tabular}
\end{table}

\section{Prompting for \abbr}
\label{sec:prompt}

In this section, we present the complete prompt formats used in the \abbr process (see \Cref{fig:method-2} for details).
\begin{tcolorbox}[colframe=black!75!white, colback=white!95!white, sharp corners=south, title=Q\&A Template, fontupper=\ttfamily]
\{Question\}

Please reason step by step, and put your final answer within \textbackslash\textbackslash boxed\{\}.
\end{tcolorbox}

\begin{tcolorbox}[colframe=black!75!white, colback=white!95!white, sharp corners=south, title=Decomposing into Thinking Units, fontupper=\ttfamily]
You will be provided with a math problem and a solution generated by a reasoning model. The model's response may contain multiple Reasoning Rounds.
One Reasoning Round is a part of the full model generation and is defined as a complete reasoning process or verification process that explicitly contains the final answer.
Your task is to carefully analyze the response and segment it into individual Reasoning Rounds. Specifically, insert ``[split]'' between every two consecutive Reasoning Rounds.

---

Problem:
\{question\}

Solution:
\{prediction\}

---

Please give the solution with ``[split]'' tags without any redundant words.
\end{tcolorbox}

\end{document}